\documentclass{article}

\usepackage{arxiv}
\PassOptionsToPackage{numbers}{natbib}

\usepackage[utf8]{inputenc}
\usepackage[T1]{fontenc}
\usepackage{hyperref}
\usepackage{url}
\usepackage{booktabs}
\usepackage{amsfonts}
\usepackage{nicefrac}
\usepackage{microtype}

\usepackage{times}
\usepackage{epsfig}
\usepackage{graphicx}
\usepackage{subfigure}
\usepackage{amsmath}
\usepackage{amssymb}
\usepackage{eucal}
\usepackage{mathtools}

\title{Solving Archaeological Puzzles}

\author{
  Niv Derech\\
  Technion, Israel\\
   \And
  Ayellet Tal \\
  Technion, Israel\\
  \And
  Ilan Shimshoni\\
  University of Haifa, Israel\\
}

\begin{document}
\maketitle
\begin{figure}[h]
\centering
\subfigure[A fresco of St. Cosmas and Damian, Serbia (26 pieces)]
{{\includegraphics[width=0.4\linewidth]{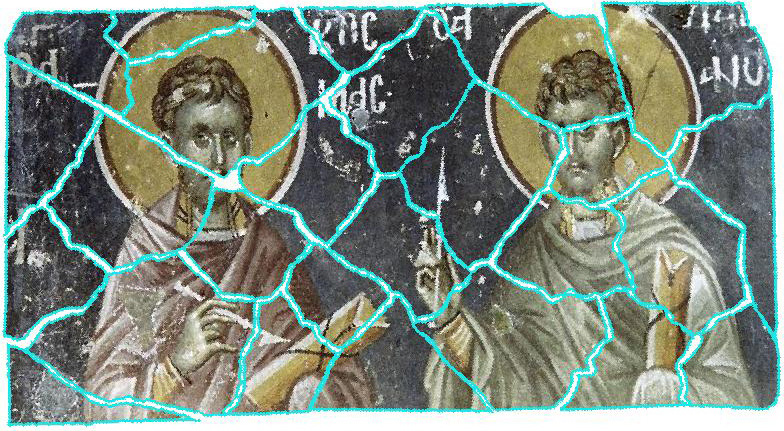}}}
\hspace{0.1in}
\subfigure[from a terracotta statue---Salamis, Cypro-Archaic period ($700$-$475$ BC).]
{\includegraphics[width=0.3\linewidth]{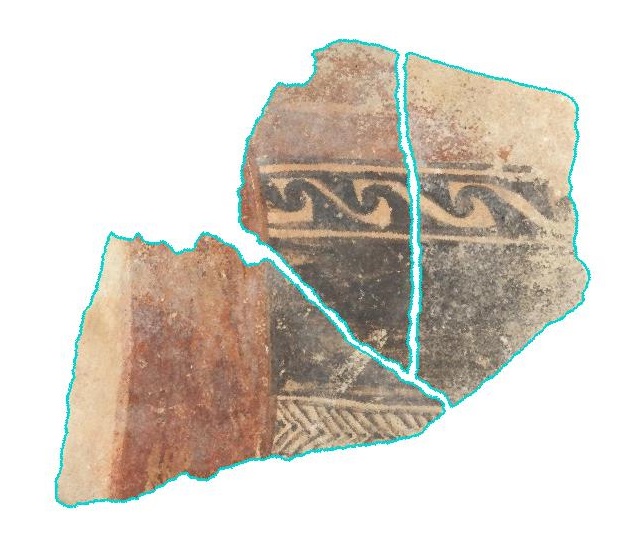}}
\caption{\label{fig:teaser}Real archaeological fragments are re-assembled by our algorithm. The borders of the fragments are marked in cyan.}
\end{figure}
\begin{abstract}
Puzzle solving is a difficult problem in its own right, even when the pieces are all square and build up a natural image.
But what if these ideal conditions do not hold?
One such application domain is archaeology, where restoring an artifact from its fragments is highly important.
From the point of view of computer vision, archaeological puzzle solving is very challenging, due to three additional difficulties: 
the fragments are of general shape;
they are abraded, especially at the boundaries (where the strongest cues for matching should exist); 
and the domain of valid transformations between the pieces is continuous.
The key contribution of this paper is a fully-automatic and general algorithm that addresses puzzle solving in this intriguing domain.
We show that our state-of-the-art approach manages to correctly reassemble dozens of broken artifacts and frescoes.
\end{abstract}

\section{Introduction}
Puzzle solving has been an intriguing problem for many years.
It has numerous application areas, such as in
shredded documents~\cite{Alpher13},
image editing~\cite{Alpher09},  
biology~\cite{Alpher04},
and archaeology~\cite{Alpher07,Alpher08,ThomasFunk:2017:WPR}.
Though the problem is NP-complete~\cite{Alpher02}, different solutions have been proposed.
The first computational solver was introduced in 1964~\cite{FreemanG64} and was able to handle a nine-piece puzzle.
Nearly five decades later, this problem is still as fascinating and intriguing.
State-of-the-art techniques mostly address natural images and are based on color matching~\cite{Alpher25,Alpher28,GurB17,Alpher30,conf/cvpr/PaikinT15,Alpher27,Alpher29,TsamouraP10,Alpher26},
on shape matching~\cite{Alpher17,Alpher16,SonMHC16}, 
or on a combination of both~\cite{Alpher19,Alpher18,Alpher21,Alpher23,Alpher22,Alpher20,ZhangL14}.

We focus in this paper on the domain of archaeology (Figure~\ref{fig:teaser}).
The vast majority of archaeological objects are discovered in a fragmentary and poor state, which
hampers the extent of archaeological research that is possible to do on them.
Much effort and time is expended in trying to manually reassemble these fragments as accurately and completely as possible.

We concentrate on archaeology not only because cultural heritage has been acknowledged worldwide as an important goal, but also because the archaeological domain exposes the limits of current computer vision techniques.
Archaeological artifacts are not “clean” and “nicely-behaved”;
rather they are broken, eroded, noisy, and ultimately extremely challenging to algorithms that analyze or reassemble them.
Therefore, from the point of view of  vision, archaeology serves as an extremely
challenging application area.

Three major differences distinguish between square-piece puzzles of natural images, which have been addressed extensively before, and images of archaeological artifacts.
These differences result from abrasion, color fading, and continuity.
Abrasion creates gaps between the pieces, which makes it harder to match adjacent fragments.
Color fading results in spurious edges.
These should be distinguished from real edges and gradients, which are highly important for matching.
Lastly, unlike square pieces for which only $16$ possible relative transformations exist between any pair of pieces, in our case the valid transformations belong to a continuous space.
This raises many questions: what is a valid transformation? how to sample the space of rigid transformations? how to compute matching scores, knowing that the transformation is imprecise due to sampling?

We propose a novel algorithm that handles these difficulties. 
It is based on four key ideas.
First, in order to address fragment abrasion, we propose to extrapolate each fragment prior to reassembly.
This reduces the continuity problem (predicting how to ``continue'' the fragment) we are facing into a matching problem.
Second, we suggest a transformation sampling method, which is based on the notion of {\em configuration space}, and is especially tailored to our problem. 
Third, at the core of any puzzle solving algorithm lies the question of what makes a good match.
We propose a new measure, which takes into account the special characteristics of the domain: the gaps between the pieces, color fading, spurious edges, the fact that the length of the matching boundary varies, and the imprecise transformations.
Finally, our placement considers not only the above scores, but also our confidence in the match, which is influenced by the uniqueness of the match and the fragment size. 

Our algorithm has been tested on real objects from the British Museum collections and on frescoes from various churches world-wide.
We demonstrate state-of-the-art results on these examples.

\noindent
{\bf Contribution.} This paper make several contributions:
First and foremost, it proposes a novel approach for solving puzzles whose pieces are of general shape, which need not necessarily match precisely due to erosion, and which have degraded unevenly over time.
The algorithm is successful due to the following components, which are required for handling archaeological artifacts, yet may find uses in other domains.
(1)~We define the notion of {\em valid transformations} and propose a method of sampling them.
(2)~We define a novel dissimilarity score that takes into account the special characteristics of the domain.
(3)~We propose not to rely solely on the dissimilarity score, but rather to measure the confidence in the match and use it during placement.

\section{Algorithm Outline}
\begin{figure}[htb]
\centering
\subfigure[Input]{
\fbox{\includegraphics[height=0.2\linewidth]{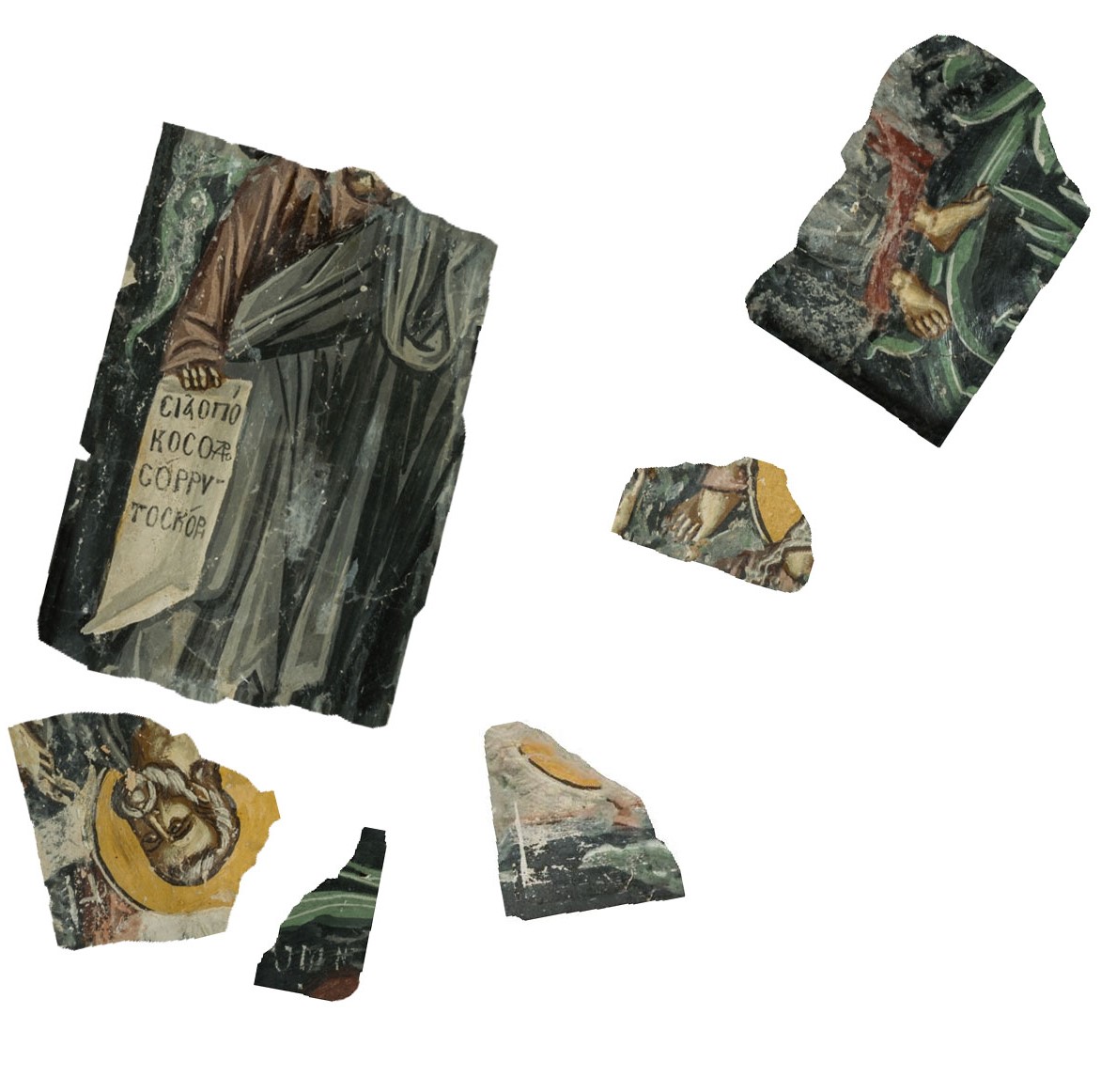}}
\label{fig:outline:input}
}
\subfigure[Fragment extrapolation ]{
\fbox{\includegraphics[height=0.2\linewidth]{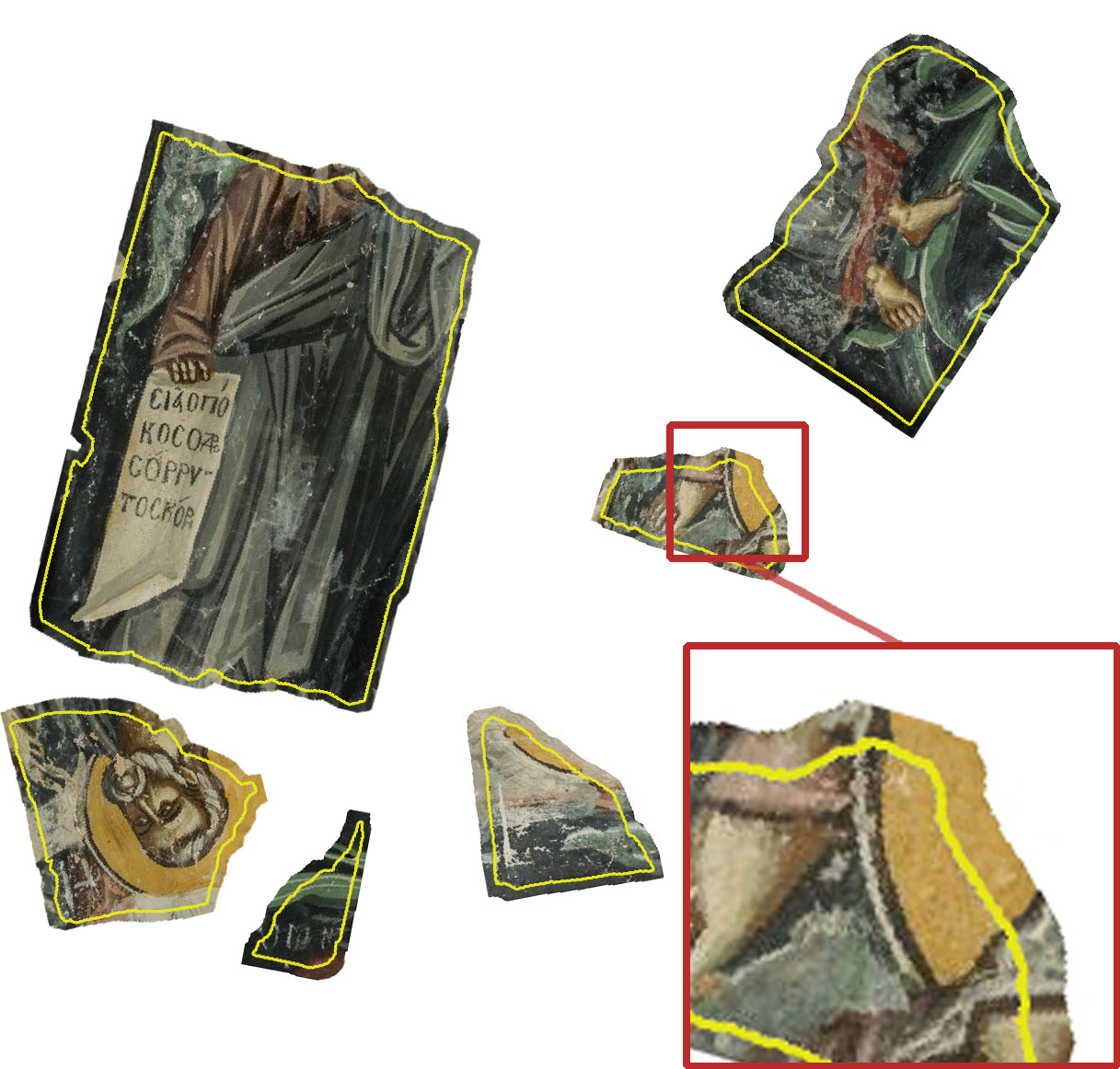}}
\label{fig:outline:extrap}
}
\subfigure[Transformation sampling]{
\fbox{\includegraphics[height=0.2\linewidth]{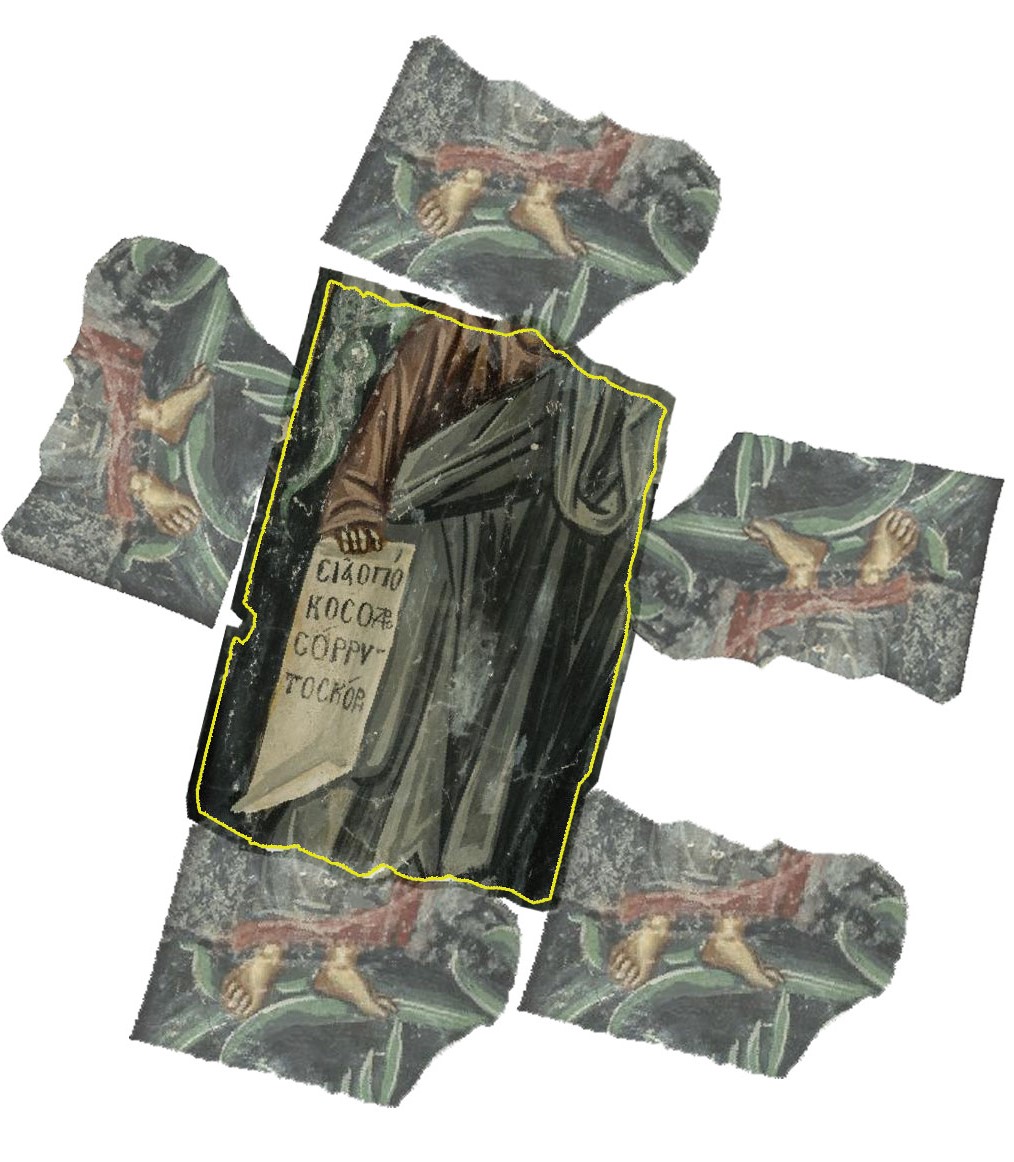}}
\label{fig:outline:diss}
}
\subfigure[Dissimilarity scores]{
\fbox{\includegraphics[height=0.2\linewidth]{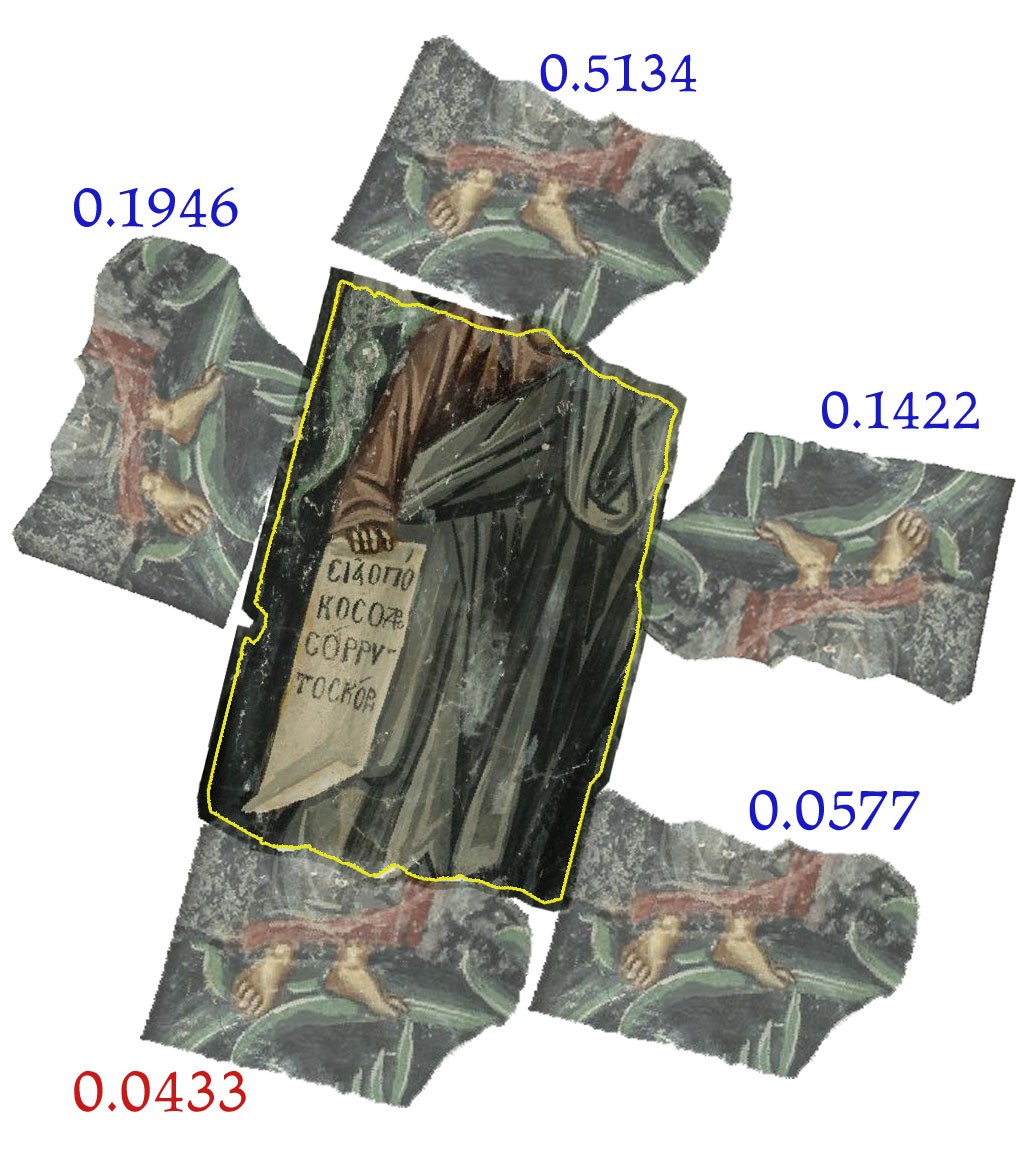}}
\label{fig:outline:firstpair}
}
\subfigure[Placement ]{
\fbox{\includegraphics[height=0.2\linewidth]{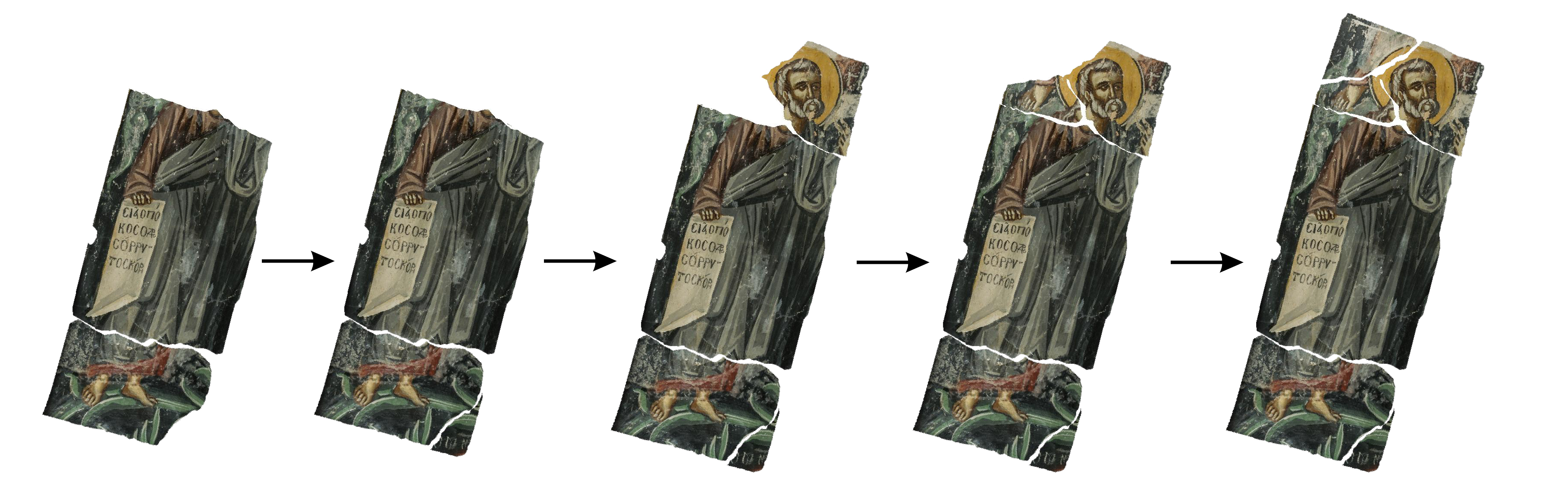}}
\label{fig:outline:placement}
}
\caption{{\bf Algorithm outline.} }
\label{fig:outline}
\end{figure}

Given $N$ images, each capturing a single fragment of an artifact, our goal is to reassemble the artifact.
In particular, each fragment should be assigned a 2D rigid transformation that indicates how to place it within the reassembly. 
Our proposed  algorithm consists of four steps, as illustrated in Figure~\ref{fig:outline}.
We briefly discuss each of the steps below and elaborate on Steps 2-4 in subsequent sections.

\noindent
{\bf 1. Fragment extrapolation.} 
Since a continuation problem is more challenging than matching, we convert our problem into a matching problem.
This is done  by extrapolating each fragment.
Formally, given a fragment, the goal is to produce a $k$-pixel wide band around it, which predicts not only the eroded region, but also the continuation to the next fragment.

Several approaches have been proposed to extrapolation~\cite{AidesAS11:fov_vid_extrap,Huang:ImageCompletion14,HuangD18,IizukaS017,PolegP12:mosaicing}.
We adopt the example-based approach of~\cite{Darabi12:ImageMelding12}, which manages to produce sharper and relatively more accurate extrapolations for our examples. 

\begin{figure}[htb]
\centering
\subfigure[]{
\includegraphics[height=0.19\linewidth]{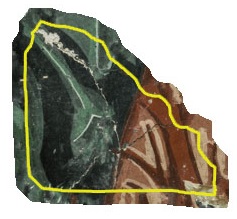}
\label{fig:extrap:ortho_14_p5}
}
\subfigure[]{
\includegraphics[height=0.19\linewidth]{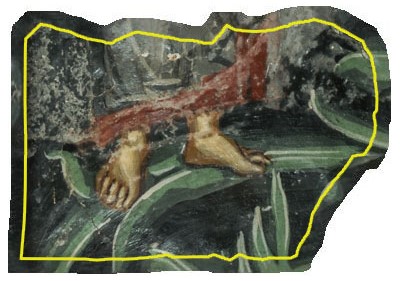}
\label{fig:extrap:ortho_24_p5_m}
}
\subfigure[]{
\includegraphics[height=0.19\linewidth]{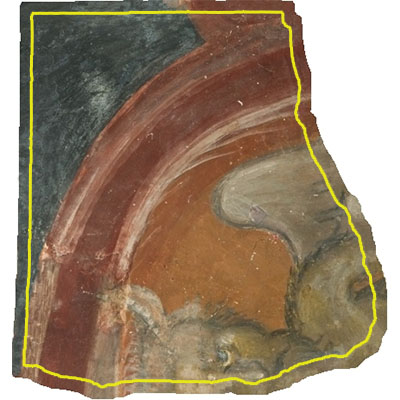}
\label{fig:extrap:ortho_34_p5_f}
}
\subfigure[]{
\includegraphics[height=0.19\linewidth]{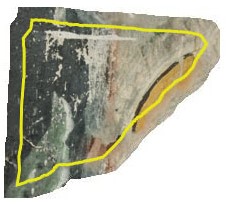}
\label{fig:extrap:ortho_24_p1_im}
}
\caption{{\bf Extrapolation results.} 
The original boundaries are marked in yellow.
Note how the patterns continue nicely in~\subref{fig:extrap:ortho_14_p5}-~\subref{fig:extrap:ortho_34_p5_f}.
However, in~\subref{fig:extrap:ortho_24_p1_im}  the yellow circle does not continue as expected, due to lack of similar patches. 
}
\label{fig:extrap}
\end{figure}

Figure~\ref{fig:extrap} shows some of our extrapolation results, where the width of the extrapolated strip depends on the fragment size ($1/6$ of the bounding box perimeter).
The results manage to continue the colors and the structure of the image in most cases.
It is interesting to note that despite of the imperfect extrapolation of Figure~\ref{fig:extrap:ortho_24_p1_im}, our algorithm manages to solve the relevant puzzle flawlessly (Figure~\ref{fig:outline}).

\vspace{0.1in}
\noindent
{\bf 2. Sampling a set of valid transformations.}
The goal is to guarantee that the transformed fragments are close to each other (so that the likelihood of adjacency is high), but do not overlap.
We therefore define a transformation between two fragments to be valid if the pixels of the original fragments do not overlap after applying it, but the extrapolated pixels of the fragments do.
We show how the set of valid transformation can be described as a configuration space problem, used in robotics~\cite{Latombe:1991:RMP:532147}.
Once the set of valid transformations is defined, we sample it  uniformly, as testing all valid transformations is infeasible.
Later, during placement, we will fine-tune the most promising transformations, to improve accuracy.
We elaborate on this step in Section~\ref{sec:sampleXform}.

\vspace{0.1in}
\noindent
{\bf 3. Finding good matches.}
We aim at computing a dissimilarity score for a given transformation between a pair of fragments.
Thanks to Step 1, the continuity problem was reduced to matching the extrapolated region of one fragment and the other original fragment, on the overlapping region.
The strongest cue, in addition to color, is that color edges from one fragment match those of the other.
However, two difficulties arise when attempting to match edges:
(1)~imprecise transformations (due to sampling) cause misalignment of the edges and (2)~degradation may yield spurious edges that should be ignored.
Our algorithm handles both problems.

A low dissimilarity value does not necessarily imply confidence in a match, since other matches with the same fragment may be as likely.
Therefore, we compute a confidence score, which predicts the likelihood of the pair of fragments to be neighbors.
The confidence in the match takes into account, in addition to the dissimilarity score, also the scores of competing matches in the same region of the fragment.
This step is explained in detail in  Section~\ref{sec:match}.

\vspace{0.1in}
\noindent
{\bf 4. Placement.}
Similarly to most previous work, our placement strategy is greedy, iteratively adding one (``best'') fragment a time.
New complications that are due to both the continuity of the space of transformations and the general shape of the fragments, need to be efficiently addressed.
Furthermore, to handle the problem of the imprecision of the transformation (due to  sampling), the selected  transformation should be refined in a small neighborhood around it.
We address all these issues in Section~\ref{sec:Placement}.

\section{Sampling valid transformations} 
\label{sec:sampleXform}
Given a pair of fragments, the goal is to find the best transformation between them.
Hence, in this section we (1)~define {\em valid} transformations; (2)~propose an efficient algorithm to validate a transformation; and (3)~since computing the dissimilarity scores for all valid transformations is intractable, as this is a continuous space, we propose an algorithm for sampling the space of valid transformations.

\begin{figure}[htb]
\centering
\includegraphics[height=0.5\linewidth]{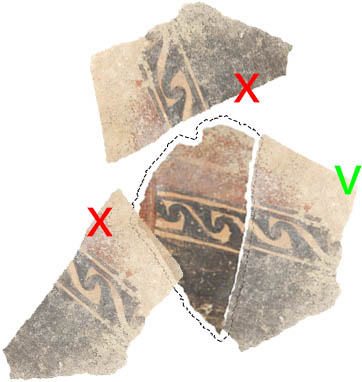}
\caption{{\bf Valid transformations.} 
The fragment must overlap the extrapolated strip, but not the original fragment 
(the extrapolation is marked with a dotted line).
Marked in green is a valid transformation, and in red are two invalid transformations---one due to intersection with the fragment and the other due to not intersecting the extrapolation.
}
\label{fig:sampling:valid}
\end{figure}

\vspace{0.1in}
\noindent
{\bf Valid Transformations.}
We are given two fragments $f_i, f_j$, and their extrapolations $f_i^+, f_j^+$.
A legal transformation is one where $f_i$ and $f_j$ do not overlap.
In addition, the archaeologists require that the two fragments should be quite close to each other.
Therefore, we define a transformation $T$ to be valid if the extrapolated fragment $f_i^+$ overlaps the transformed fragment $T(f_j)$, but the original fragment $f_i$ does not overlap it.
Formally, $T$ should satisfy the following constraint, as illustrated in Figure~\ref{fig:sampling:valid}:
\begin{equation}
\label{eq:sampling:valid set}
T(f_j)\cap f_i^+\neq \emptyset\ \wedge\ T(f_j) \cap f_i = \emptyset.
\end{equation}

\vspace{0.1in}
\noindent
{\bf Efficient transformation validation.}
Verifying the validity (Equation~\eqref{eq:sampling:valid set}) of every sampled transformation is very expensive.
The key idea of bypassing this procedure is to borrow the concept of a {\em configuration space} from the field of robotics~\cite{Latombe:1991:RMP:532147}.
This reduces the problem of checking for overlap between fragments to the problem of checking whether a point is inside a polygon. 

Briefly, given a translating and rotating robot $\mathcal{R}$ and an obstacle $\mathcal{P}$, the configuration space is a 3-dimensional space $\mathbb{R}^2\times [0:2\pi)$.
A point $p=(x,y,\theta)$ in the configuration space corresponds to a certain placement of the robot in the 2D plane after rotating the robot by $\theta$
and translating it by $(x,y)$.
The configuration space is defined as the set of points such that the corresponding placement of $\mathcal{R}$ after rotation by $\theta$ does  intersect $\mathcal{P}$. 
This can be efficiently computed by the Minkowski sum operator $\oplus$ as:
\begin{equation}
\label{eq:configuration_space}
\mathcal{C}(\mathcal{P},\mathcal{R},\theta)=\mathcal{P}\oplus -\mathcal{R}(0,0,\theta),\quad
\end{equation}
where 
$\mathcal{R}(0,0,\theta)$ is the robot's location after rotation by $\theta$  and no translation. Once the Minkowski sum has been computed as a polygon, we can easily test whether a point is in the configuration space (i.e., inside the polygon).
The configuration space is computed for each $\theta$ independently.

The free space (free of intersection) is the complementary of Equation~\eqref{eq:configuration_space}, i.e.,
\begin{equation}
\label{eq:free_space}
\mathcal{F}(\mathcal{P},\mathcal{R},\theta)=\overline{\mathcal{C}(\mathcal{P},\mathcal{R},\theta)}.
\end{equation}

In our case, we wish to convert Equation~\eqref{eq:sampling:valid set} to Equations~\eqref{eq:configuration_space}-\eqref{eq:free_space}.
Let fragment $f_j$ be the robot and $f_i$ \& $f_i^+$ be the obstacles.
We require that $T(f_j)$ will not intersect $f_i$ (free space), but will intersect $f_i^+$ (configuration space).
Therefore, we can formulate our validation operator $\mathcal{V}$ as
\begin{equation}
\label{eq:validate}
\mathcal{V}(f_i,f_j,\theta)=\mathcal{F}(f_i,f_j,\theta) \wedge \mathcal{C}(f_i^+,f_j,\theta).
\end{equation}

Equation~\eqref{eq:validate} must be computed also when the roles of $f_i$ and $f_j$ are switched, for the inverse transformation.
The transformation is valid when both conditions hold.

\vspace{0.1in}
\noindent
{\bf Sampling algorithm.}
Once we have an efficient method for validly testing, sampling can be performed in three simple steps, as follows. 
First, $N_r$ rotations (angles) are sampled uniformly. 
Then, for each sampled rotation, we construct two translational configuration spaces, one for $f_i,f_j$ and the other for $f_i^+,f_j$, resulting in two polygons.
The two polygons are sampled uniformly at resolution $R_t$, and transformations that satisfy Equation~\eqref{eq:validate} are maintained.

In our experiments we set $N_r=80$ and $R_t$ to $5$ pixels in each direction.
This suffices to find a high-score transformation in the neighborhood of the correct transformation.
Later on, during  placement,  local refinements of the recovered transformations are performed.

\noindent
{\bf Complexity:} 
Configuration space construction is linear in the size of the fragments (assuming convexity); this is performed $N_r$ times.
The point-in-polygon procedure is linear in the size of the polygon and performed for every sample.

An alternative approach is to find a globally optimal transformation using a branch-and-bound strategy~\cite{conf/cvpr/LitmanKBA15}. 
In our case, this procedure is impractical, both due to its complexity and due to the difficulty to incorporate Equation~\eqref{eq:validate}, as well as our dissimilarity function, into the framework.

\section{Finding good matches} 
\label{sec:match}

A good dissimilarity score is a fundamental component in every jigsaw puzzle solver. 
It predicts the likelihood of two aligned fragments to be neighbors.
However, other fragments may also have good scores and compete for the same region of the fragment.
We therefore add another measure, which indicates our confidence in a match.
Below, we describe our dissimilarity and confidence scores.

\subsection{Dissimilarity score} 
\label{sec:dissimilarity}

Given two pieces, $f_i, f_j$  and a transformation $T$, we are seeking a dissimilarity measure $d(f_i,f_j,T)$ that predicts the likelihood of $f_i$ and $f_j$ to be neighbors, after transforming $f_j$ by $T$. 
What would make a good match?
First, the patterns and the colors of $f_i$ should continue seamlessly to $T(f_j)$.
This is especially important near the gradients, as edge continuation is a strong hint for correct reassembly.
Second, the length of the match should be considered; the longer the match, the higher the probability that the match is correct and not random. 

Our algorithm is based on three key ideas.
First, we view the extrapolation as a prediction of the continuation of the fragment.
Therefore, it suffices to compare the overlap of the vicinity of the boundary of one fragment with the extrapolation of the other.
Second, we look for the continuation of image gradients between the pieces.
However, since due to erosion, there are superfluous gradients, we discard these gradients using knowledge about our domain.
Third, even a slight error in the transformation between the fragments causes the color gradients to be misaligned.
Since our transformations are sampled, slight misalignment must be ignored when computing the dissimilarity score.
We elaborate on the realization of these ideas hereafter.

Let $\Psi_{i,j}^T=\partial f_i\cap T(f_j^+)$ be the curve defined as the overlap of the boundary of $f_i$ with the extrapolation of the transformed $f_j$.
This curve defines a set of pixels both in  $f_i$ and in  $f_j^+$, which should be compared.

To accommodate the first idea, we compute the color difference  for the pixels on $\Psi_{i,j}^T$ , as follows:
\begin{equation}
\label{eq:color_diss}
d_c(f_i,f_j^+,T)=\sum\limits_{p\in\Psi_{i,j}^T}D(f_i(p),T(f_j^+)(p)),
\end{equation}
where $D(u,v)$ is the $L_1$ distance between the colors of pixel $u$ and $v$ in the LAB color space.

In line with idea (2), we would like to disregard the gradients that are due to erosion.
We make use of the fact that archaeological artifacts use only a few colors (that are culture-dependent).
We therefore generate a {\em color palette} to represent the original colors~\cite{Finkelstein15:palette}.
A small set of colors is selected, which represents the full range of colors in the image.
This is done using a variant of the $k$-means algorithm, where $k$ is the number of colors.
As $k$ is unknown, we compute palettes for different values of $k$ ($k=3,4,6,8,10$).
The principle is that the smaller the value of $k$ in which the gradient appears, the more important this gradient is. 

Before explaining how the different palettes are utilized, recall that due to the inaccuracy of the transformations, the gradients are slightly misaligned. 
To jointly realize ideas~(2)-(3), we build a weighted histogram of the gradient directions. In this histogram, slight misalignments do not affect the score.
Furthermore, each gradient is weighted by $1/k$.
Thus, the weight of a gradient is high when it is repeated in many palettes and when it appears early on, in palettes consisting of a small number of colors.

\begin{figure}
    \centering
    \subfigure[Original]{\includegraphics[width=0.24\linewidth]{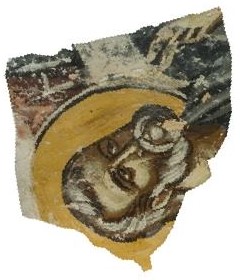}}
    \subfigure[$k=3$]{\includegraphics[width=0.24\linewidth]{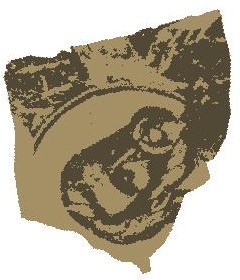}}
    \subfigure[$k=4$]{\includegraphics[width=0.24\linewidth]{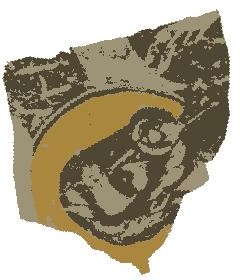}}
    \subfigure[$k=6$]{\includegraphics[width=0.24\linewidth]{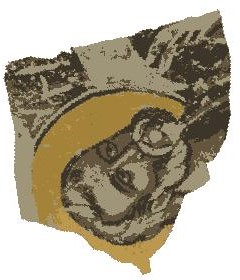}}
    \subfigure[$k=8$]{\includegraphics[width=0.24\linewidth]{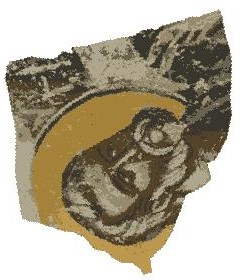}}
    \subfigure[$k=10$]{\includegraphics[width=0.24\linewidth]{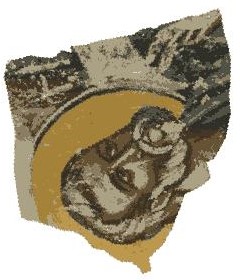}}
    \subfigure[Ours]{\includegraphics[width=0.24\linewidth]{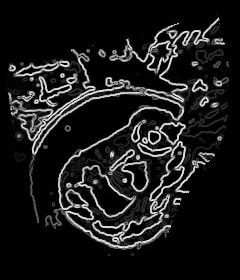}}
    \subfigure[Sobel filter]{\includegraphics[width=0.24\linewidth]{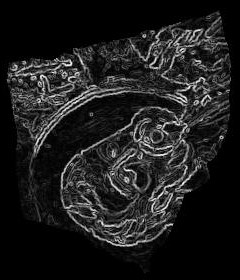}}
    \caption{{\bf Palettes for gradient computation}. 
   As the number of colors ($k$) in the palette grows, more details become visible. 
   Gradients that appear early on and remain consistent across many palettes, are more important and hence are weighted stronger.}
    \label{fig:pallette}
\end{figure}

Figure~\ref{fig:pallette} demonstrates this method. 
As we add more colors to the palette, the image is further segmented, showing more details.
Yet, the edges that are due to erosion still do not appear.
Furthermore, the important edges appear in all palettes, hence they are stronger.
Contrary to classical edge detection, these are the only apparent edges.

Given two histograms of gradients on $\Psi_{i,j}^T$, $h_i,h_{j^+}$, for $f_i,f_{j^+}$, respectively.
These histograms are smoothed with a Gaussian kernel, in order to deal with matching gradient directions that might fall in adjacent bins.
The difference between the histograms is computed as follows:
\begin{equation}
\label{eq:hist_diff}
d_h(f_i,f_j^+,T) = \sum\limits_d |h_i(d)-h_{j^+}(d)|(1+\ln(a(d))).
\end{equation}
Inspired by~\cite{Ghafurian:2017:CER}, we give more weight to larger bin differences;
the latter is weighted by $1+\ln(a(d))$, where
\begin{equation}
a(d)=\frac{\max(h_i(d),h_{j^+}(d))}{\min(h_i(d),h_{j^+}(d))}.
\end{equation}

\noindent{\bf Putting it all together.}
The combined dissimilarity score takes into account both Equations~\eqref{eq:color_diss} and ~\eqref{eq:hist_diff}, as well as the length of the intersecting curve $\Psi_{i,j}^T$.
It is defined as:
\begin{equation}\label{eq:diss}
d(f_i,f_j^+,T)=\frac{d_h(f_i,f_j^+,T) + d_c(f_i,f_j^+,T)}{|\Psi_{i,j}^T|^\alpha},
\end{equation}
where  the parameter $\alpha$ controls the importance of this length ($\alpha=2$ is used in all our examples).
The dissimilarity score, which indicates the quality of the match, is low when the color difference and the histogram difference are small, and the intersection curve is long. 

\subsection{Confidence} 
\label{sec:Confidence}
The goal is to evaluate the confidence that a small dissimilarity between two fragments indeed indicates adjacency. 
Given a fragment, it may be the case that many candidates could match it in the same place with very good scores.
In such a case, we cannot be confident that a specific match is the correct one. 
Conversely, our confidence should be high when a match is unique.

We are given a pair of fragments $f_i,f_j$ and a transformation $T$ for which we want to compute the confidence.
The key idea is to compare $T$'s dissimilarity with those of the {\em competing transformations} and to use this comparison to determine $T$'s uniqueness.
To do that, we shall define what the competing transformations are.
We will define them below, noting that they can be associated either with $f_i$ \& $f_j$ or between $f_i$ and some other fragment.
In the first case, we have to disregard  $T$'s nearby transformations, as they would naturally give similar dissimilarity scores.

\vspace{0.1in}
\noindent
{\bf Disregarding nearby transformations.}
To disregard nearby transformations, we have to define the distance between transformations.
One could consider to measure the distance in transformation space, but since we are interested in matches, we propose a distance that suits our problem.

Given two valid transformations $T_1,T_2$ applied to $f_j$, we consider $\Psi_{i,j}^{T_1}, \Psi_{i,j}^{T_2}$, the two curves of intersection within $f_i$, after applying the transformations to $f_j$.
(Note that the curve might be fragmented.)
We compute the {\em Intersection over Union (IoU)} between the curves and define it as the distance between transformations $T_1$ and $T_2$:
\begin{equation}
    \label{eq:dist_transform}
    XfSim_{i,j}(T_1,T_2) = \frac{\Psi_{i,j}^{T_1}\cap \Psi_{i,j}^{T_2}}{\Psi_{i,j}^{T_1}\cup \Psi_{i,j}^{T_2}}.
\end{equation}
The larger the overlap, the larger this value.
Hence, if this value is larger than a predefined threshold $\gamma_H$, the transformations are considered close to each other, and therefore are not considered competitors  ($\gamma_H=0.85$ in our experiments).
The same condition is also tested when the roles of $f_i$ and $f_j$ are reversed.
Hence, $T_1$ and $T_2$ are regarded as competitors if
\begin{equation}
XfSim_{i,j}(T_1,T_2) < \gamma_H \lor XfSim_{j,i}(T_1^{-1},T_2^{-1}) < \gamma_H.
\label{eq:close}    
\end{equation}

\vspace{0.1in}
\noindent
{\bf Confidence computation.}
The key idea is to look for the most competitive match on $\Psi_{i,j}$ within $f_i$.
If the dissimilarity of this competitive transformation is much worse than the score of~ $T$, then our confidence in $T$ is high.

Specifically, we constructively define the set of {\em competitive transformations} as follows: 
We apply Equation~\eqref{eq:dist_transform} to the curves of every sampled transformation of the pair $(f_i, f_m), \forall m$.
Every transformation $S$ for which $XfSim_{i,m}(T,S) > \gamma_L$ is competitive. 
Intuitively, this condition implies that the overlap between the matches, defined by transformations $S$ and $T$ (of some fragment with fragment $f_i$), is high.
We note that for $m=j$, $T$ and $S$ should also not be nearby (Equation~\eqref{eq:close}).

Now that a set of competitive transformation is defined, we proceed to define our confidence $C(f_i,f_j,T)$ in transformation $T$.
We choose from this set the best match with $f_i$, $competitor_i(T)$ i.e., the one with the lowest dissimilarity score.
If the best competitor and $T$  have similar scores, we are not confident that $T$ is indeed the correct match.
However, if their dissimilarities to $f_i$ differ greatly, our confidence in the match is high.
This is expressed in the following definition of the confidence:
\begin{equation*}
 \hat{C}(f_i,f_j,T)=1-\frac{d(f_i,f_j^+,T)}{d(f_i, f_m^+,competitor_i(T))}.
\end{equation*}

To further increase our confidence in a match, we recall the concept of {\em best buddies} \cite{Alpher27}.
Intuitively, two pieces are best buddies if they both agree that the other piece is their most likely neighbor.
Thus, we would like to consider also the inverse transformation.
However, since our transformations are sampled, the inverse of $T$ might not have been sampled.
To solve this, we define our confidence in matching two fragments $f_i$ and $f_j$ to be high if there exist two transformations, $T_1$ and $T_2$, such that
(1) $T_1$ transforms $f_j$ about $f_i$ with large confidence and 
(2) $T_2$, which is ``almost'' inverse to $T_1$, transforms $f_i$ about $f_j$ and also has high confidence.
This is expressed as follows: 
\begin{equation}\label{eq:conf}
C(f_i,f_j,T_1,T_2)=\bigg(\frac{\hat{C}(f_i,f_j,T_1) + \hat{C}(f_j,f_i,T_2)}{2}\bigg),
\end{equation}
where 
$$XfSim(T_1,T_2^{-1})\geq\gamma_H \wedge
XfSim(T_1^{-1},T_2)\geq\gamma_H.$$

\section{Placement}
\label{sec:Placement}
The last phase of the algorithm is the reassembly the artifact from its fragments. 
The complexity of the problem is extremely high, as given $N$ fragments, there are  $\binom{N}{2}$ pairs, and for each of them an infinite number of transformations (from which we sampled a finite set).
Our goal is to find $N$ transformations that will result in the correct reassembly.

We follow the most common strategy in puzzle solving---greedy reassembly~\cite{Alpher28,GurB17,Alpher30,conf/cvpr/PaikinT15,Alpher27}, which iteratively selects the fragments to be placed.
At every iteration, rather than looking for a piece that best matches a single previously-placed piece, we look for a piece that maximizes the confidence with respect to multiple-placed pieces.
Differently from previous work, this should be done when the transformations are sampled from a continuous space, rather than having only $4(/16)$ possible placements.

For each pair of fragments 
we construct a set of the best (dissimilarity-wise) non-competing transformations between them.
For each transformation in this set, the confidence is computed, as explained in Section~\ref{sec:match}.

The placement is initialized by choosing the pair of fragments with the maximal confidence  $C(f_i,f_j,T_1,T_2)$, for some transformation pair $T_1,T_2$ (Equation~\eqref{eq:conf}).
Then, the following stages are iterated.
\begin{enumerate}
    \item
    {\bf Discarding illegal placements:} Illegal (physically-impossible) transformations between placed and unplaced fragments are discarded.
\item
    {\bf Updating the confidence:} The confidence values of candidate transformations, which match multiple-placed fragments, are re-calculated.
\item
    {\bf Placement:} The next best piece is placed, after refining its corresponding transformation.
\end{enumerate}
We elaborate on the above steps hereafter.
Let $P$ be the set of the already-placed fragments at step $k$, where each fragment $f_i \in P$ is associated with transformation $T_i$.

In Step 1 we discard illegal transformations, which would place a fragment $f_j$ on top of  a fragment from $P$. 
To efficiently verify whether $T(f_j)$ is legal, we use the configuration space that was already calculated during transformation sampling (Section~\ref{sec:sampleXform}).
For $T$ to be legal, we require that it does not belong to the configuration space of $\mathcal{C}(T_i(f_i),f_j)$ for all $f_i \in P$.

How can this be determined efficiently?
We have only computed $\mathcal{C}(f_i,f_j)$ before, whereas we seek to know $\mathcal{C}(T_i(f_i),f_j)$.
The trick we use is noting that the condition that  $T \notin \mathcal{C}(T_i(f_i),f_j)$ is equivalent to the condition that $T_i^{-1} T \notin \mathcal{C}(f_i,f_j)$.
But, $\mathcal{C}(f_i,f_j)$ was already computed. 

Once a partial placement already exists, the confidence values  of any given transformation might no longer be valid and therefore must be re-calculated.
This is the goal of Step 2.
This requires us to first detect the neighbors of a transformed fragment, and then calculate the dissimilarity \& confidence values with the already-placed pieces.

Note that a given transformation is valid with regard to multiple-placed fragments, if it is valid (Equation~\ref{eq:sampling:valid set}) for at least one placed fragment and legal for the rest of them.
This means that we can use the previously-computed configuration spaces, as done in Step 1. 

\begin{figure*}[htb]
\centering
\begin{tabular}{c}
\includegraphics[height=0.15\linewidth]{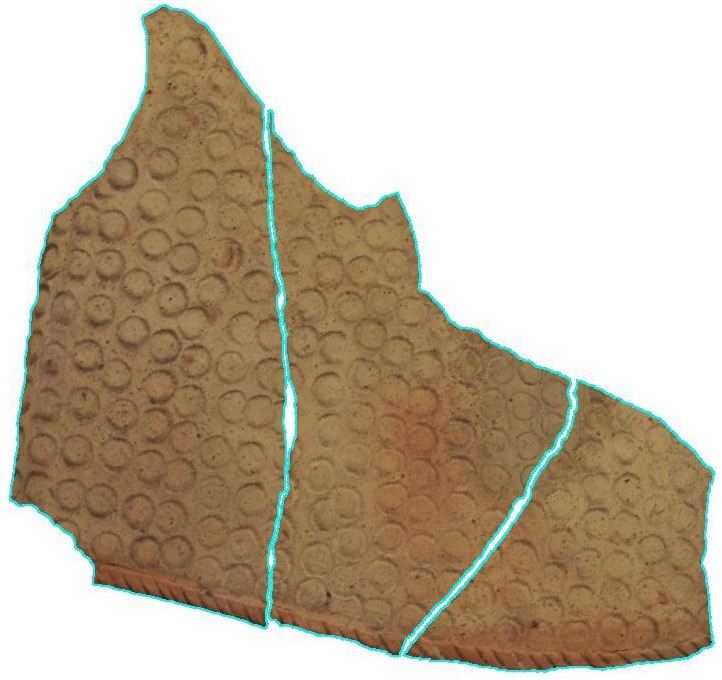}
\includegraphics[height=0.15\linewidth]{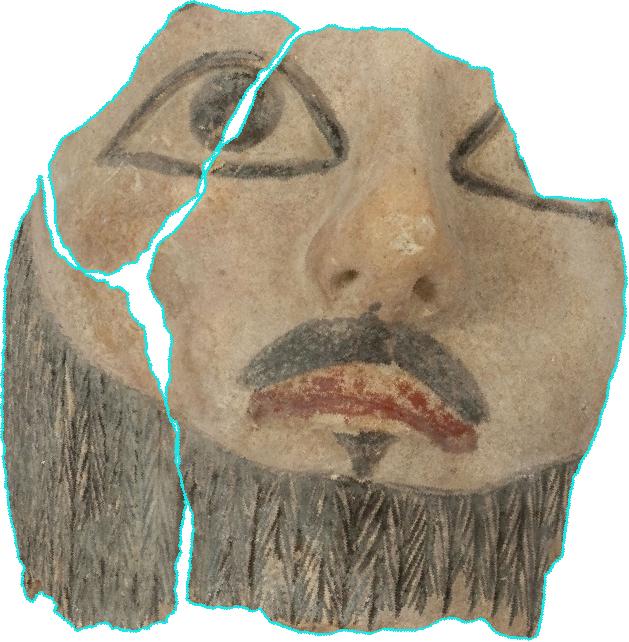}
\includegraphics[height=0.15\linewidth]{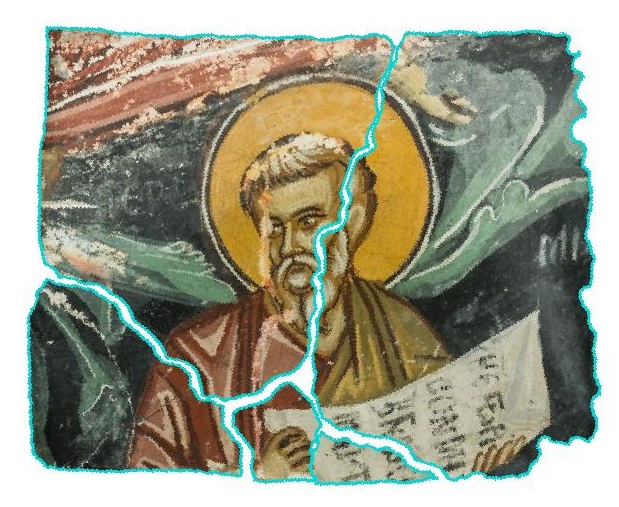}
\includegraphics[height=0.15\linewidth]{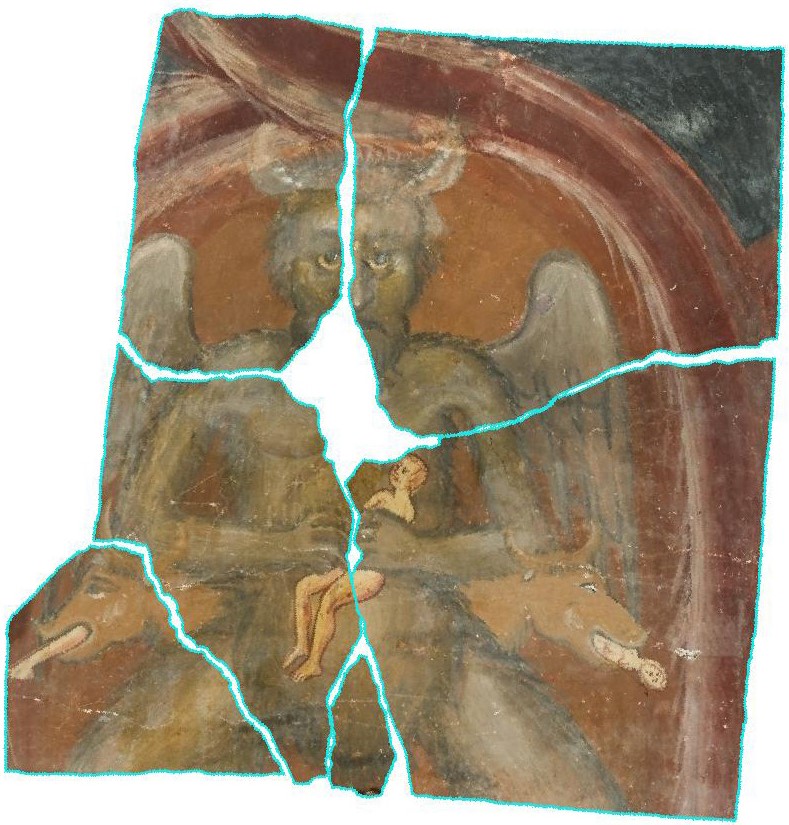}
\includegraphics[height=0.135\linewidth]{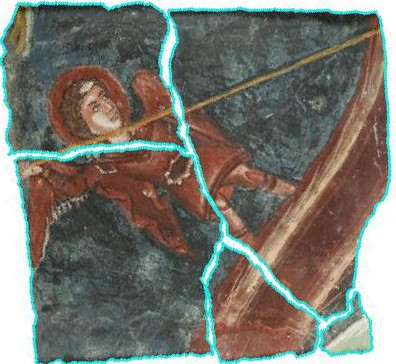}
\includegraphics[height=0.135\linewidth]{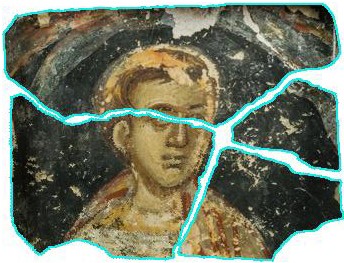}
\end{tabular}
\caption{{\bf Reassembly of real artifact.}
The left two examples are of 3D statues from Salamis, while the remaining four are frescoes from the church of Christ Antiphonitis in Cyprus.
The fragments are faded, stained, scratched, and with other apparent  degradation artifacts.
Furthermore, they do not accurately match due to erosion or missing pieces.
Our algorithm manages to re-assemble them perfectly.}
\label{fig:real}
\end{figure*}

The dissimilarity and the confidence should now be re-computed.
The key idea is to consider all the already-placed fragments as a single (virtual) fragment, $f_P$ and compute these values as usual (Section~\ref{sec:match}).

Finally, in Step 3, the fragment $f_j$ with the highest confidence is placed using the corresponding transformation $T_j$.
The latter is now refined, by densely re-sampling the space of transformations in a small neighborhood around  $T_j$, in order to address the inaccuracy of the transformation due to sampling.
Then, the dissimilarities of the new sampled transformations to $f_P$ are calculated.

The final transformation $\tilde{T_j}$ is the one that minimizes:
\begin{equation}
\begin{split}
F(T)&=d(f_P,f_j^+,T)+d(f_j,f_P^+,T)
\\
\tilde{T_j} &= \arg\min\limits_{T\in \mathcal{N}(T_j)}F(T),\\
\end{split}
\end{equation}
where $\mathcal{N}(T_j)$ is the set of sampled transformations in the neighborhood of $T_j$.

We note that the final assembly may change, depending on the initial transformation sampling.
This might result in missing some correct matches.
To solve this, we could either increase the sampling resolution or perform the algorithm multiple times on randomly-rotated fragments and choose the best assembly.
As the second solution is cheaper, we do it and run the algorithm five times.
Then, our algorithm automatically chooses the best reassembly, which is measured by its overall (sum) confidence. 

\section{Results}

This section describes the results of our algorithm, when applied to real archaeological artifacts.
In some cases, the fragments are naturally broken, while in others, they were generated to emulate the way artifacts have been broken.

\vspace{0.05in}
\noindent
{\bf Qualitative results.}
Figure~\ref{fig:real} shows the reassembly of some real artifacts.
Some of the examples include images of fragments of 3D statues, whereas others are images of frescoes.
These examples have eroded fragment boundaries, missing pieces, and faded colors.
Some have very few colors, while others have more; moreover, the same colors appear in different regions of the artifact.
It can be seen that our algorithm re-assembled all the artifacts flawlessly.

\begin{figure}[htb]
    \centering
    \subfigure[Dry mud]{
    \includegraphics[width=0.47\linewidth]{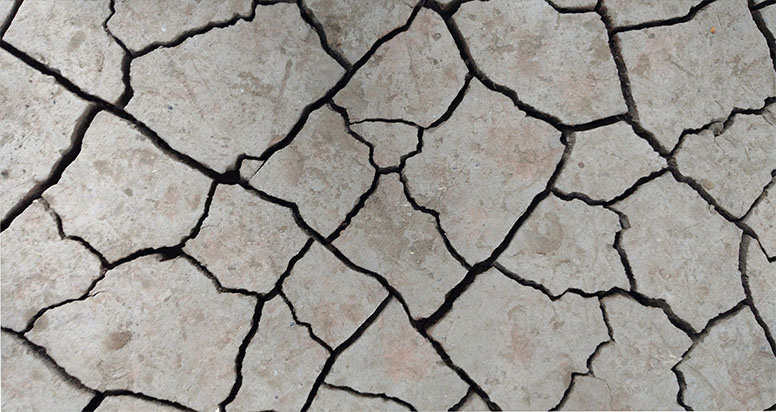}
    \label{fig:fragmentation:drymud}
    }
    \subfigure[Corresponding fragmentation]{
    \includegraphics[width=0.47\linewidth]{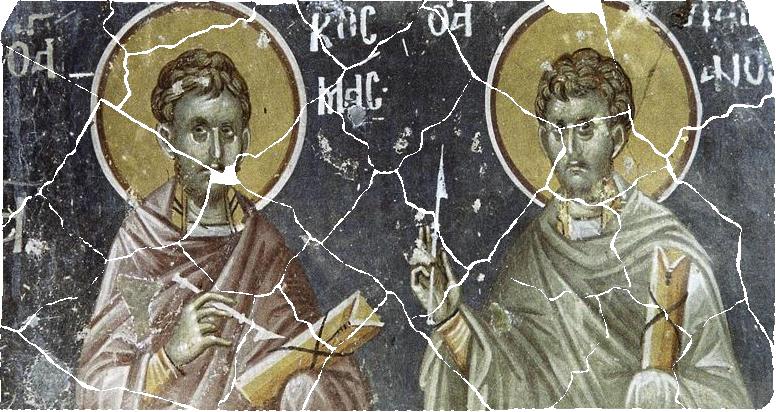}
    \label{fig:fragmentation:brokenorig}
    }
    \caption{{\bf Fragment generation.}
    To generate ``natural'' fragments, we break an artifact/fresco based on a fragmentation of dry mud. }
    \label{fig:fragmentation}
\end{figure}

Since we could not obtain artifacts of this type with more fragments, we generated some challenging examples in a novel way, which simulates natural fragmentation, as follows.
Given an image of a real artifact/fresco, we created its fragmentation according to a pattern of dry mud, as shown in Figure~\ref{fig:fragmentation}.
Dry mud not only breaks in a natural manner, but also generates gaps between the pieces.

\begin{figure*}[htb]
\centering
\subfigure[{\em Apse of Sant Climent de Taüll}, $12^{th}$ Century (18 pieces)]
{{\includegraphics[height=0.2\linewidth]{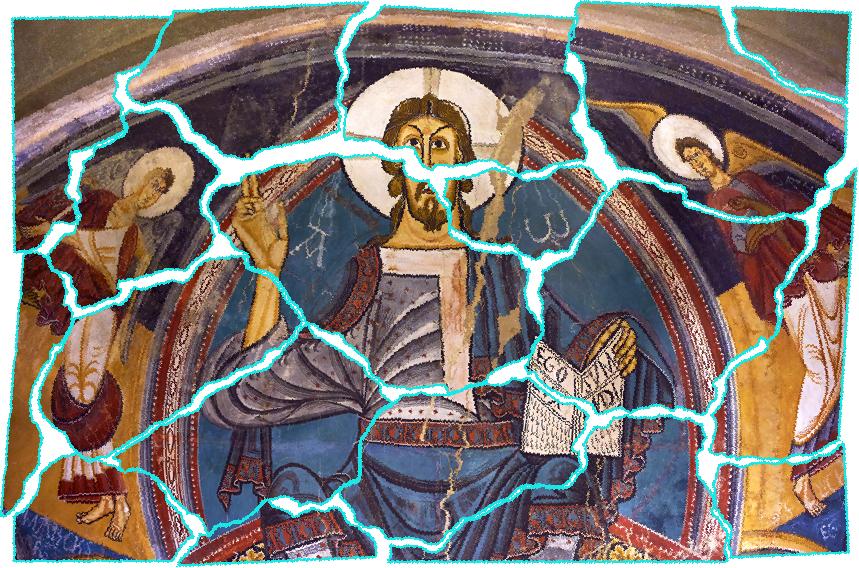}}}
\subfigure[from Arles in France, $20$-$70$ BC (18 pieces)]
{{\includegraphics[height=0.2\linewidth]{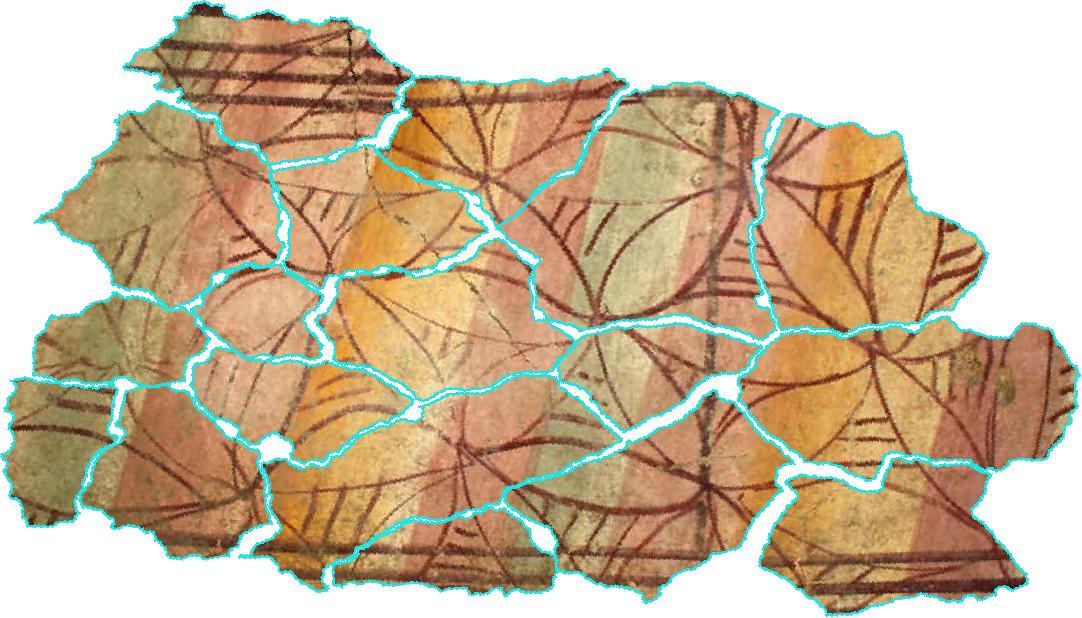}}}
\subfigure[from the {\em Panagia tou Araka} Church in Cyprus (18 pieces)]
{{\includegraphics[height=0.2\linewidth]{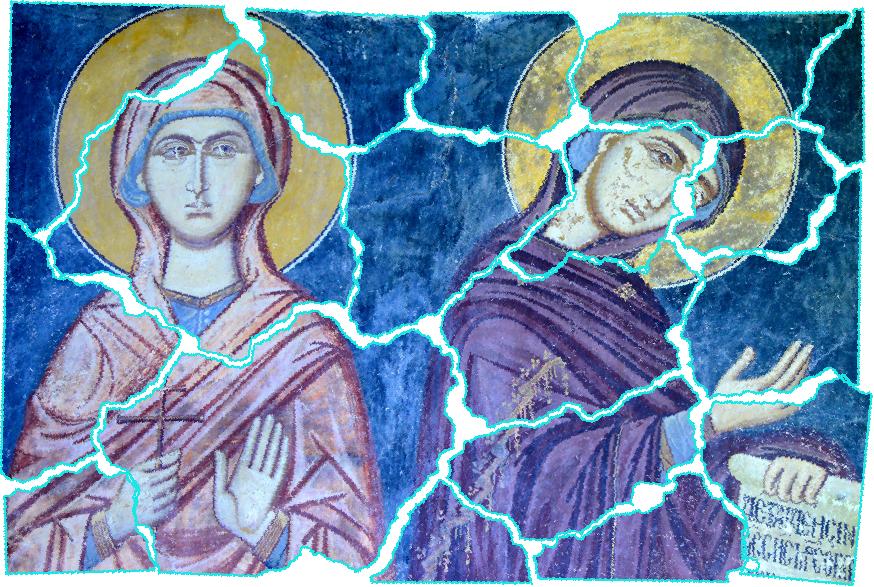}}}
\subfigure[from the {\em Panagia tou Araka} Church in Cyprus (26 pieces)]
{{\includegraphics[height=0.2\linewidth]{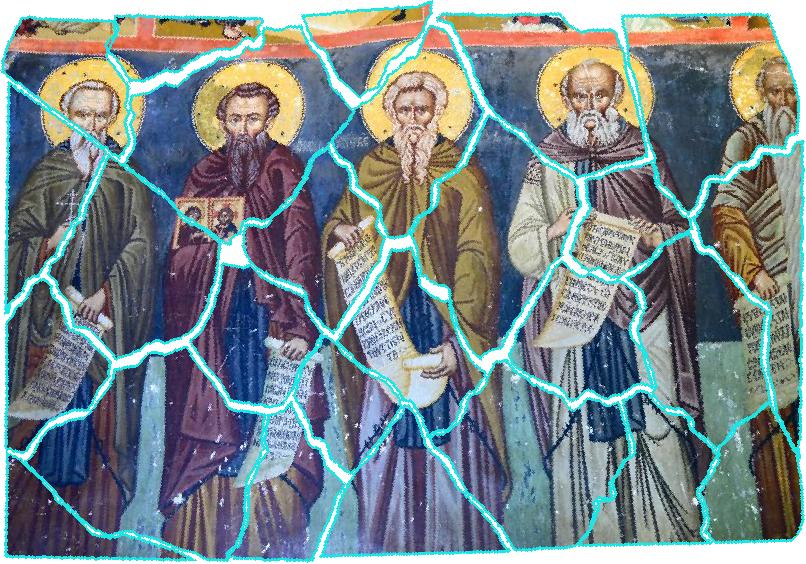}}}
\subfigure[from the {\em Agios Nikolaos tis Stegis} church in Cyprus (26 pieces)]
{{\includegraphics[height=0.2\linewidth]{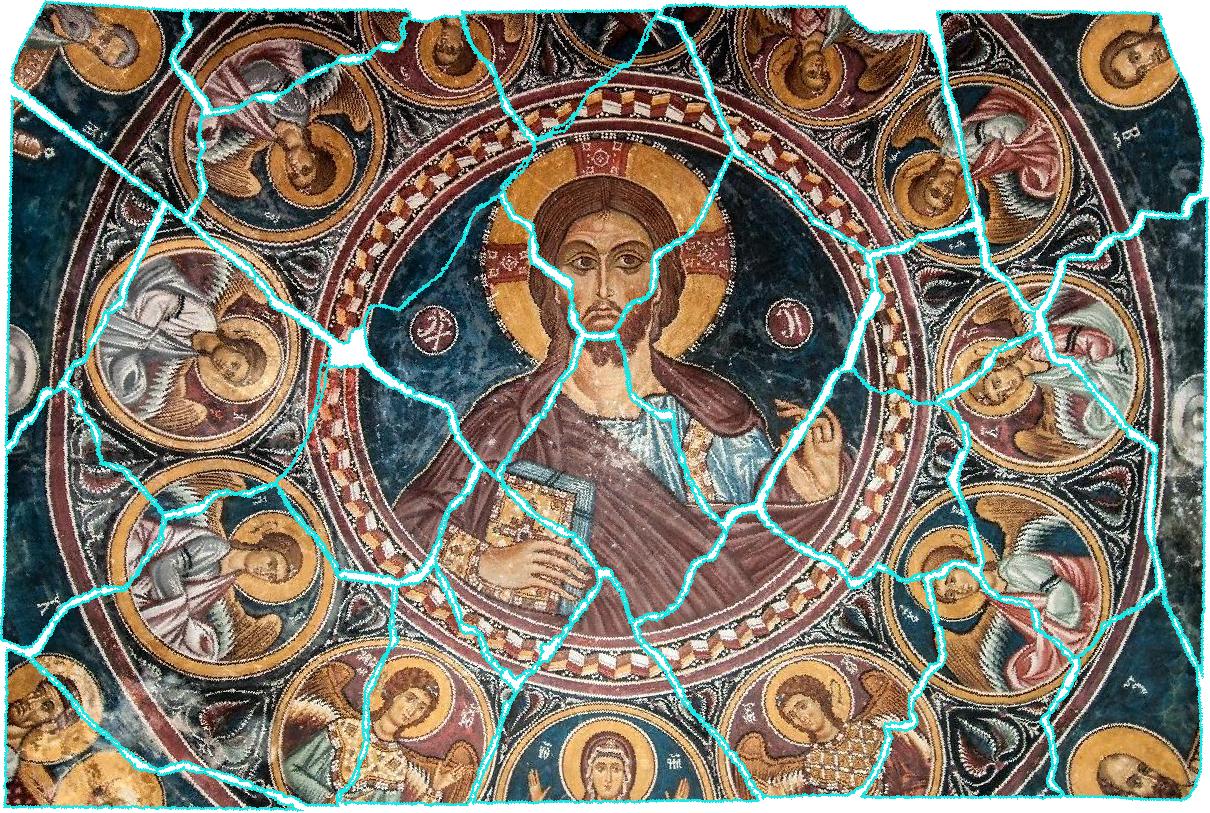}}}
\subfigure[from the {\em Stravos tou Agiasmati} Church in Cyprus (36 pieces)]
{{\includegraphics[height=0.2\linewidth]{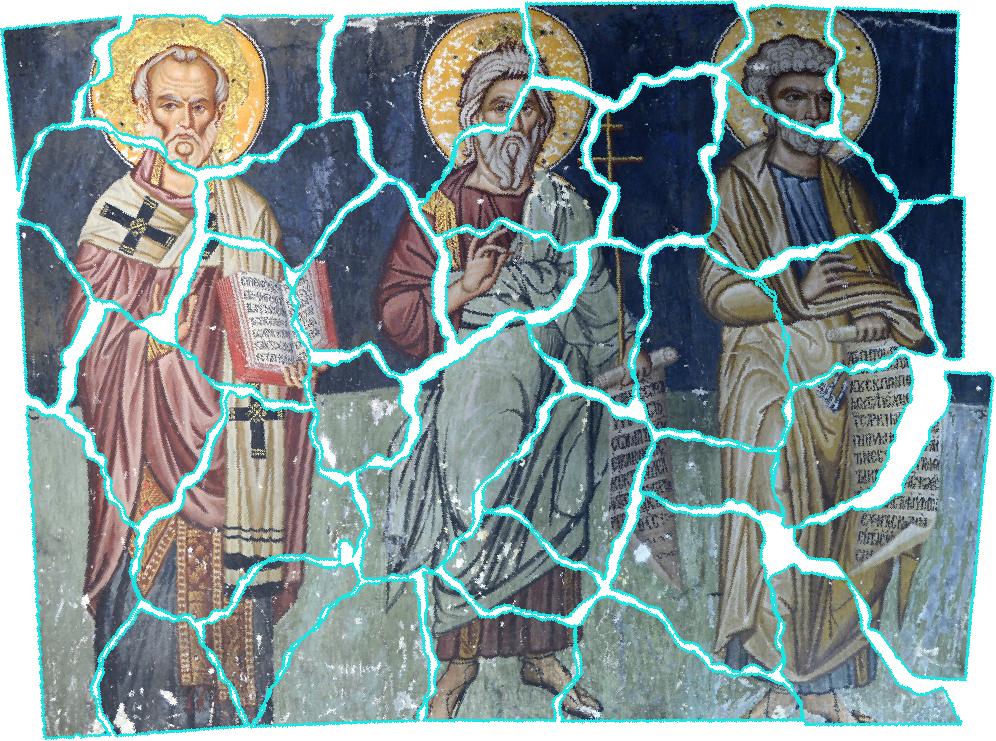}}}
\caption{{\bf Qualitative results.} 
The frescoes were broken into fragments using a variety of dry-mud patterns, and each fragment was randomly rotated.
The geometric partition varies, as well as the patterns and the colors.
Some have many repeating patterns, which makes these examples more difficult to solve;
some have only a few colors that occupy large regions, whereas others have a larger variety of colors.
Still, our algorithm managed to reassemble these examples flawlessly.
}
\label{fig:mud}
\end{figure*}

Figure~\ref{fig:mud} demonstrates our re-assembly for several such examples.
Both the number of fragments and the shape of the patterns vary.
Color fading is due to the artifacts themselves.
Our algorithm is shown to perfectly reassemble the vast majority of these artifacts, while missing only a few matches in the others.

\vspace{0.05in}
\noindent
{\bf Quantitative results.}
The most common measure used for square-pieces puzzles of natural images~\cite{Alpher09,Alpher30,conf/cvpr/PaikinT15,Alpher27,Alpher29} is the {\em  average neighbor}. 
It reports on the fraction of correct pairwise adjacencies. 

Since in our case the transformation space is continuous (rather than having only 16 possible transformations), we suggest a modification of the popular neighbor measure.
It is defined as the fraction of the correct adjacent fragment pairs, where correct is defined as satisfying the condition that the intersection fraction between the relative transformation between the pair and the relative ground truth transformation applied to each of the fragments, is larger than a given threshold ($\frac{2}{3}$ in our setting).

Table~\ref{table:measures} reports on the results of this adjusted measure. 
The results are averaged over all examples with the specified number of fragments. 
It is evident from the table that the results are almost perfect.
This is so, since in general-shaped puzzles we require that not only the pairs be adjacent, but also that the transformation is relatively accurate.

\begin{table}[htb]
\centering
\begin{tabular}{||c|c|c||}
\hline
\# pieces& \# artifacts& neighbor\\
\hline
$\leq 10$ & 8& 100\%\\
\hline
$11$-$20$ &13&92.60\%\\
\hline
$21$-$30$&4&98.50\%\\
\hline
$31$-$40$&6&93.81\%\\
\hline
\end{tabular}
\caption{{\bf Quantitative results} }
\label{table:measures}
\end{table}

\begin{figure}[tb]
    \centering
    \subfigure[Original image ($26$ fragments)]{\includegraphics[height=0.28\linewidth]{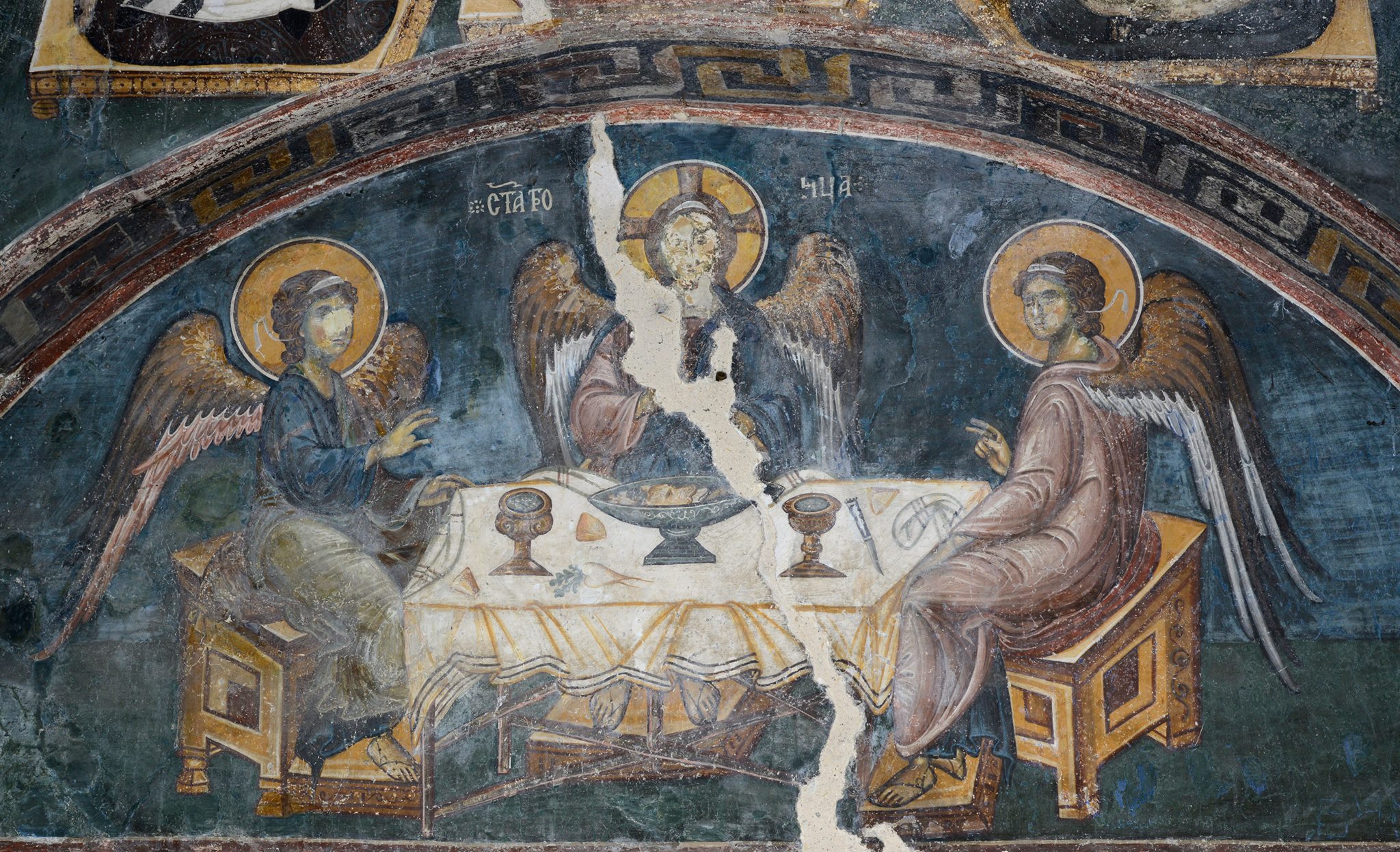}}
    \subfigure[assembly result]{\includegraphics[height=0.28\linewidth]{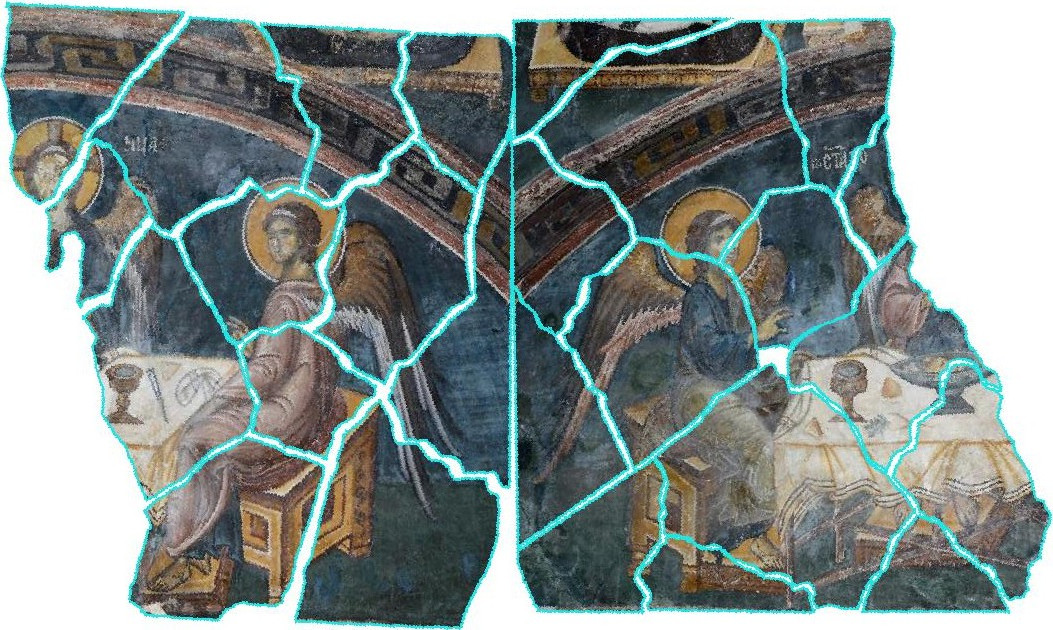}}
    \caption{{\bf Limitation.} The algorithm successfully reassembled the pieces on both sides of the large white gap, but failed to find the correct matches across the gap. 
    }
    \label{fig:limitation}
\end{figure}

\noindent
{\bf Limitations.}
Figure~\ref{fig:limitation} shows an example for which our algorithm fails to perfectly reassemble a fresco.
Since this fresco contains a large gap in the middle, neighbors on both sides of the gap are not matched.
Rather, matches that continue the patterns were selected.
Nevertheless, the neighbor measure is still quite high ($0.94$).

\vspace{0.05in}
\noindent
{\bf Closely-related works.}
The most closely related works to ours are~\cite{ZhangL14,ThomasFunk:2017:WPR,Huang:2013:MGT:2508363.2508373}, which assume general-shaped pieces.
However, in~\cite{ZhangL14} the images are natural and there are no gaps between the pieces.
\cite{Huang:2013:MGT:2508363.2508373} focuses on the gaps, but here again, the images are natural.
Moreover, the user provides an approximate initial placement.
Our algorithm successfully solves~\cite{Huang:2013:MGT:2508363.2508373}'s examples, but without  initial placements.

In~\cite{ThomasFunk:2017:WPR} a highly challenging archaeological example is solved.
The focus is on matching the geometry (boundaries), as this puzzle is uncolored.
Conversely, we focus on matching the colors; the geometry is not directly addressed.

\section{Conclusion}
\label{sec:conclusion}
This paper introduced a novel approach for solving puzzles whose pieces are of general shape, whose boundaries are abraded, and whose colors are damaged.
To deal with these challenges, our algorithm is based on four key ideas.

First, prior to reassembly, each fragment is extrapolated, predicting how it should be continued by its adjacent piece.
This reduces the problem of continuity of the fragments to the problem of matching them.
Second, valid transformations between two fragments are defined as transformations for which one fragment does not intersect the other, but does intersect its extrapolation.
Using the concept of {\em configuration space}, borrowed from robotics, allows us to compute the valid transformations efficiently.
Third, a dissimilarity score is developed, which addresses the problems arising from degradation, as well as the imprecision of the transformations due to sampling.
Fourth, the confidence in the match is computed, by choosing the more relevant competitors and comparing their dissimilarities.
Both of these values are computed using a novel definition of the distance between transformations.

Our algorithm was evaluated on a variety of archaeological broken artifacts and frescoes.
It was shown, both qualitatively and quantitatively,  to faithfully reassemble them.

\paragraph{Acknowledgements:}
\label{sec:Acknowledgements}
This research was partially supported by the GRAVITATE project under EU2020-REFLECTIVE-7-2014 Research and Innovation Action, grant no. 665155 and the Israel Science Foundation (ISF) 1083/18.

\newpage

\bibliographystyle{splncs}
\bibliography{references}
\end{document}